\documentclass[11pt]{article}

\usepackage[final]{acl}

\usepackage{times}
\usepackage{latexsym}

\usepackage[T1]{fontenc}

\usepackage[utf8]{inputenc}

\usepackage{microtype}

\usepackage{inconsolata}

\usepackage{graphicx}

\usepackage{amsmath}   
\usepackage{amssymb}   
\usepackage{bm}        
\usepackage{textcomp}
\usepackage{booktabs}
\usepackage{multirow}
\usepackage{xcolor}     
\usepackage[dvipsnames]{xcolor}
\usepackage[most]{tcolorbox}  
\usepackage{colortbl}

\definecolor{qianlan}{HTML}{c2d1e5}
\definecolor{yellow}{HTML}{ffc64b}

\title{Towards Proactive Personalization through Profile Customization for Individual Users in Dialogues}

\author{
 \textbf{Xiaotian Zhang\textsuperscript{1,2}\thanks{Equal contribution.}} \
 \textbf{Yuan Wang\textsuperscript{1,2}\footnotemark[1]} \ 
 \textbf{Ruizhe Chen\textsuperscript{1,2}\footnotemark[1]} \\
 \textbf{Zeya Wang\textsuperscript{1}} \
 \textbf{Runchen Hou\textsuperscript{1}} \
 \textbf{Zuozhu Liu\textsuperscript{1,2}\thanks{Corresponding author.}}
\\
 \textsuperscript{1}Zhejiang University \
 \textsuperscript{2}Zhejiang Key Laboratory of Medical Imaging Artificial Intelligence
\\
}

\begin{document}
\maketitle
\begin{abstract}
The deployment of Large Language Models (LLMs) in interactive systems necessitates a deep alignment with the nuanced and dynamic preferences of individual users. Current alignment techniques predominantly address universal human values or static, single-turn preferences, thereby failing to address the critical needs of long-term personalization and the initial user cold-start problem. To bridge this gap, we propose PersonalAgent, a novel user-centric lifelong agent designed to continuously infer and adapt to user preferences. PersonalAgent constructs and dynamically refines a unified user profile by decomposing dialogues into single-turn interactions, framing preference inference as a sequential decision-making task. Experiments show that PersonalAgent achieves superior performance over strong prompt-based and policy optimization baselines, not only in idealized but also in noisy conversational contexts, while preserving cross-session preference consistency. Furthermore, human evaluation confirms that PersonalAgent excels at capturing user preferences naturally and coherently. Our findings underscore the importance of lifelong personalization for developing more inclusive and adaptive conversational agents. Our code is available \href{https://github.com/Monncyann/PersonalAgent.git}{here}.
\end{abstract}

\section{Introduction}

With the rapid advancement of Large Language Models (LLMs) in executing complex language tasks~\citep{li2023open,gpt4}, ensuring that their outputs remain aligned with human values and preferences has become increasingly critical~\citep{houben2022inspect,ji2023ai}. Previous alignment methodologies have predominantly focused on adherence to broad and universal human preferences, such as helpfulness and harmlessness~\citep{shen2023large}. While these principles have enabled LLMs to exhibit socially acceptable behavior across a wide user base, they often overlook the nuanced requirements of individual users who expect alignment with their implicit preferences during the interaction~\citep{wang2023mint,do}. The capacity of LLMs to accommodate the diverse needs, goals, and interaction styles of individual users, especially by proactively learning the implicit preferences that frequently arise in everyday conversations, is crucial yet under-explored for enhancing the user experience in conversational agents and boosting inclusivity in user-agent interactions.

Meanwhile, prior methods typically focus on alignment at the single-turn level, lacking mechanisms for cross-turn or even cross-session personalization. This limits the agent’s ability to maintain long-term consistency with user preferences~\citep{pad,psoups}. The core challenge arises from the inherently dynamic and evolving nature of personalization. In extended interactions, users continuously reveal preference information, which is not always directly applicable to the current request. However, effective personalization requires agents to proactively infer and adapt to user-specific attributes, retaining them over time to allow long-term alignment~\citep{do,aloe}. Moreover, existing methods for aligning to user requests assume that the agent already possesses relevant information~\cite{personaagent}. However, in real-world scenarios, the agent often encounters the user cold-start problem, with no prior user information available. Accordingly, we characterize a personal agent as follows:

\textit{\textbf{A user-centric lifelong personal agent should proactively infer user preferences and maintain a unified memory to ensure long-term consistency.}}

\begin{figure*}[t]
\centering
  \includegraphics[width=0.98\textwidth]{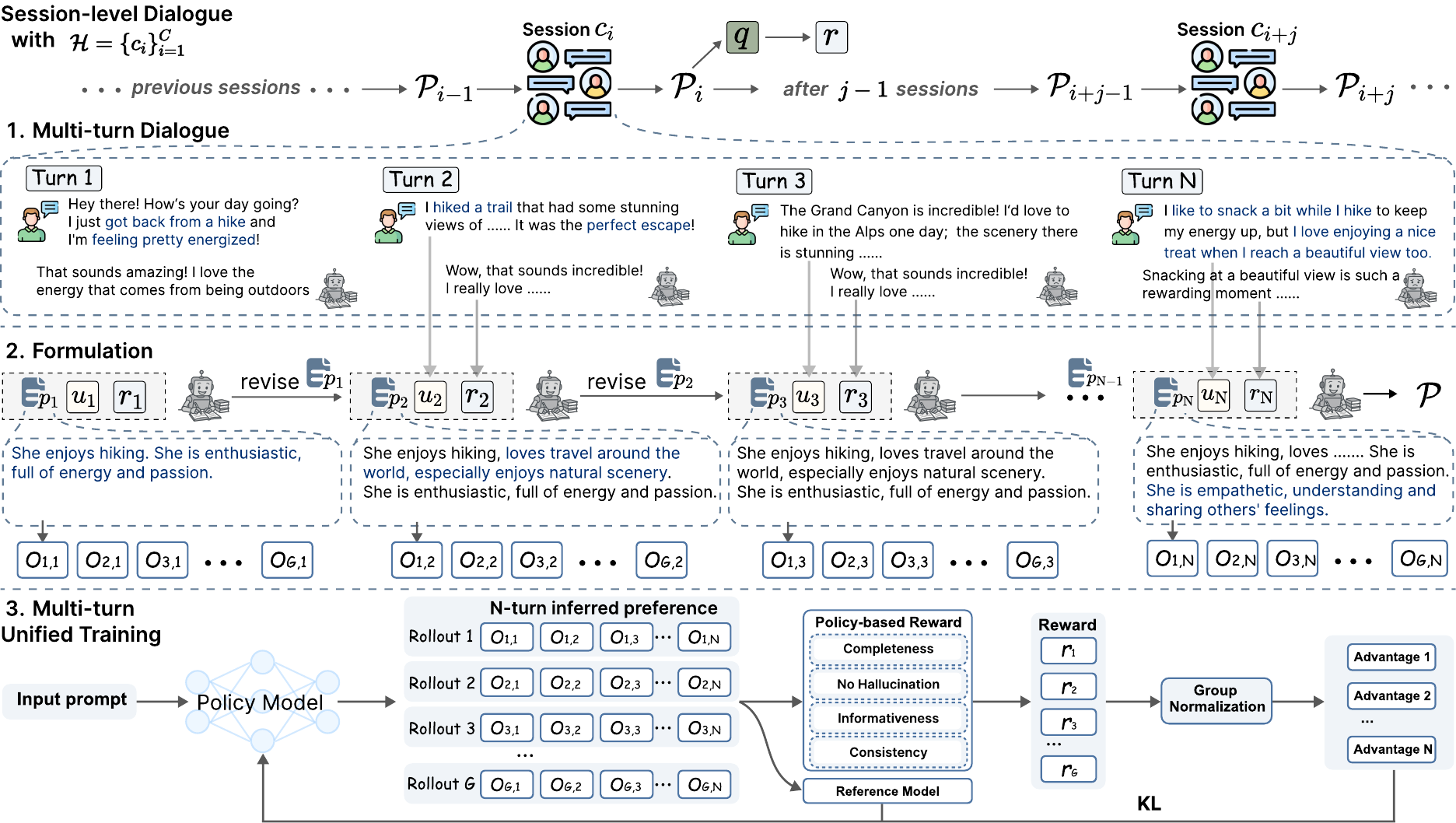}
  \caption{PersonalAgent is inspired by the way humans communicate with others. Rather than feeding the entire conversation history $\mathcal{H}$ as input, it learns multi-turn dialogues $\bm{c}$ turn by turn and processes them iteratively, recording relevant information in a user profile $\mathcal{P}$. Finally, the agent leverages the profile $\mathcal{P}$ stored across sessions to determine whether further querying is needed before generating a response $\bm{r}$ for the user request $\bm{q}$.}
  \label{fig:formulation}
\end{figure*}

In this paper, we introduce PersonalAgent, which aims to model multi-turn conversations in a manner consistent with human intuition while maintaining long-term consistency. To achieve this, we first emulate the human memory process in conversation by decomposing multi-turn dialogues into single-turn units for memory modeling. As demonstrated in Figure~\ref{fig:formulation}, each turn outputs the user preferences conveyed in the current dialogue and feeds them as input to the next turn for further refinement. This strategy incrementally processes preference inference over long texts (formulated as a multi-turn Markov Decision Process), jointly optimizes multi-turn rewards, and ultimately refines an independent user profile, thereby enabling accurate inference while ensuring long-term consistency. In addition, motivated by the lack of prior work on user cold-start scenarios, we curate and construct the ALOE-Unseen dataset to benchmark agents’ ability to proactively query users for better alignment. 

Experimental results demonstrate that PersonalAgent significantly outperforms prompt-based methods and policy optimization methods in identifying user preferences during conversation. When irrelevant dialogues are inserted during testing, the performance of these traditional methods drops substantially, whereas PersonalAgent still surpasses agent baselines that are specifically equipped with memory mechanisms. This demonstrates that PersonalAgent not only infers preferences accurately within the dialog but also maintains long-term consistency as the conversation evolves. Further analysis reveals that modeling the multi-turn dialog as a sequence of decomposed rounds enables the agent to adapt to personalization in a more natural and coherent manner, thereby achieving cross-session personalized alignment. In addition, we investigate different training strategies (Base, SFT, and RL) under the same paradigm, showing that a policy-based judge is better suited to capture the dynamics of multi-turn dialogue evolution. Finally, we conduct human annotation and long-term alignment evaluations to ensure the reliability of our results.

Our major contributions are threefold:
\begin{itemize}
    \item We decompose personalization in long-context interactions into intuitive turn-level segments and formulate it as a multi-turn Markov Decision Process (MDP), which allows unified optimization to capture and adapt to personalized preferences across turns.
    \item We maintain a lifelong profile for each individual user in session-level dialogues to ensure long-term alignment with their diverse personalized preferences.
    \item We curate and construct the ALOE-Unseen dataset, which is specifically designed to address the critical user cold-start scenario. Experiments across multiple themes and settings further demonstrate the superior performance of PersonalAgent.
\end{itemize}

\section{Related Works}

\noindent \textbf{Personalized Alignment.} Previous efforts to align LLMs with human preferences have largely relied on policy-based methods~\citep{modpo,morlhf,diffpo}, such as Reinforcement Learning from Human Feedback (RLHF)~\citep{rlhf} and Direct Preference Optimization (DPO)~\citep{dpo}. Although these approaches enable natural, human-preference-consistent instruction following, they are limited by the reliance on coarse-grained population-level alignment, which limits the model's ability to address individual user needs. Moving toward personalization, some works~\citep{map,pad,personajudge} allow users to explicitly specify the degree of alignment across single or multiple dimensions, thereby achieving more personalized objectives. However, such approaches often overlook the rich variability in individual preferences, constraining their ability to scale toward fine-grained, user-specific alignment. Recently, personalized systems such as PersonaAgent~\citep{personaagent} employ system prompts as mediators, integrating memory and action modules. Nevertheless, they still fall short in modeling multi-turn interactions and thus fail to effectively capture latent preferences that emerge over the course of interaction.

\noindent \textbf{User-Centric Personalization.} By defining role-based profiles for LLMs, previous work has enabled user analysis that fosters more natural and sophisticated personalized responses~\citep{secom,personajudge}. Personalization workflows such as profile-augmented generation (PAG)~\citep{pag} and reinforcement learning for personalized alignment (RLPA)~\citep{teaching} introduce a weak-parametric approach to personalization by integrating external user-specific data into model outputs. However, these methods mainly focus on how and what to align, while \textit{overlooking the fundamental question of whether alignment is feasible}. The work of \citet{whose} is most similar to ours, which applies abductive reasoning to preference data in order to infer users' underlying needs and interests. However, the reliance on binary preference data limits the scalability to the diverse and fine-grained spectrum of personalization, resulting in the constraint to achieve proactive personalization.

\section{Proactive Personalization}
In this section, we first formulate the multi-turn dialogue scenario (\S~\ref{sec:task formulation}), and then present the process of dynamically constructing user profiles (\S~\ref{sec:User-centric Design}). To this end, we provide a detailed description of how user preferences are inferred turn by turn (\S~\ref{sec:Preference inference}), culminating in the realization of proactive personalization (\S~\ref{sec:Response generation}). Finally, we illustrate the concrete implementation (\S~\ref{sec:Practical implementations}).

\subsection{Task Formulation}
\label{sec:task formulation}

Conversations involve dynamic interactions between users and agents, as well as extensive exchanges between the agent and the inferred user profile. At each interaction turn, the agent must communicate with the user to collect information and infer their intent while dynamically updating the user profile before generating a response to the user’s request. Let $\mathcal{H} = \{\bm{c}_i\}_{i=1}^C$ denote the conversation history between the user and the agent, which includes $C$ sessions. $\bm{c}_i = \{\bm{t}_n\}_{n=1}^{T_{i}}$ represents the $i$-th session that consists of $T_{i}$ sequential user-agent interaction turns, with each turn $\bm{t}_n=(u_n,r_n)$ including a user request $u_n$ and the corresponding response from the agent $r_n$. Denote the user-centric personalization system as $f_P$, and the response generation model as $f_{\text{LLM}}$. As shown in Figure~\ref{fig:formulation}, the overall research framework can be formalized as:
(1) \textit{Profile construction}: construct a user profile $\mathcal{P}$ using conversation history $\mathcal{H}$; $\mathcal{P}$ is learned and refined in the conversation at the turn-level, each interaction turn $t_n$ corresponds to a brief inferred profile $p_n$, with $\mathcal{P}=\sum_{n=1}^T p_n$. Then for a session-level user profile, it is initialized as $\mathcal{P}_{\text{old}}$ at the beginning of a dialogue and evolves to $\mathcal{P}_{\text{new}}$ at the end;
(2) \textit{Preference inference}: given a target user request $q$ and a user profile $\mathcal{P}$, query preferences $\{p\in\mathcal{P}\} \leftarrow f_p(q, \mathcal{P})$ that are relevant to the user request, and determines whether the current profile is sufficient to align the response with the given request $q$;
(3) \textit{Response generation}: the agent is permitted to proactively elicit extra information $p^*$ from the user to ensure better alignment. The final response is generated as $r=f_{\text{LLM}}(q, p^*, \{p\in\mathcal{P}\})$.

\subsection{User-centric Design}
\label{sec:User-centric Design}
Analogous to real-world interactions, people do not continuously revisit the entire dialogue history during a conversation; instead, they rely on impressions to carry the interaction forward. We construct a dedicated profile $\mathcal{P}=\sum_{n=1}^T p_n$ for each individual user to help the agent instantiate this abstract memory. The profile template is constructed based on the \textit{LMSYS-Chat-1M} dataset~\citep{lmsys}, which consists of one million real interactions between users and 25 state-of-the-art language models across a wide range of topics. We categorize user preferences into 11 major categories, which are further divided into over 300 subcategories, aiming to provide a comprehensive multi-dimensional description of each user. This enables the construction of highly personalized profiles that can also adapt to evolving user needs, thereby capturing the dynamic nature of user preferences. The specific categories are illustrated in Figure~\ref{fig:profile}, and Appendix~\ref{app:profile details} provides a detailed description of the construction process.

As mentioned in \S \ref{sec:task formulation}, each interaction turn $t_n$ corresponds to some inferred attributes of profile $p_n$, while each session $c_{i}$ corresponds to an aggregated profile $\mathcal{P}_{i}$ inferred jointly from all preceding sessions. It can be solved in parallel:
\begin{equation}
\label{eq:turn level}
\mathcal{P}_{i}\left\{
\begin{aligned}
p_1 &= \arg \max_q \pi(q \mid u_1) \\
p_2 &= \arg \max_q \pi(q \mid p_1, u_2) \\
& \vdots \\
p_{n+1} &= \arg \max_q \pi(q \mid p_n, u_{n+1})
\end{aligned}
\right.
\end{equation}
This design enables rapid, turn-level personalization updates while maintaining long-term consistency with the user across sessions.

\subsection{Preference inference}
\label{sec:Preference inference}
To maintain the user profile $\mathcal{P}$ according to Eq.~\ref{eq:turn level}, we discard redundant historical information and optimize using only the inferred attributes $p$ from each turn. Turn-level personalized alignment can be formulated as a multi-turn Markov Decision Process (MDP)~\citep{teaching}, defined by the tuple $(\mathcal{S}, \mathcal{A}, \mathcal{T}, \mathcal{R}, T)$, where the state space $\mathcal{S}$ consists of the current user message $u$ and inferred attributes of profile so far (i.e., $s_t = (u_t, p_{1:t-1})$). The action $a_t$ corresponds to the inferred attribute $p_t$ at turn $t$. $\mathcal{T}$ is the transition kernel, which is deterministic, that given the state $s_t = (u_t, p_{1:t-1})$ and action $a_t = p_t$, the next state is:

\begin{figure}[t]
  \includegraphics[width=1\columnwidth]{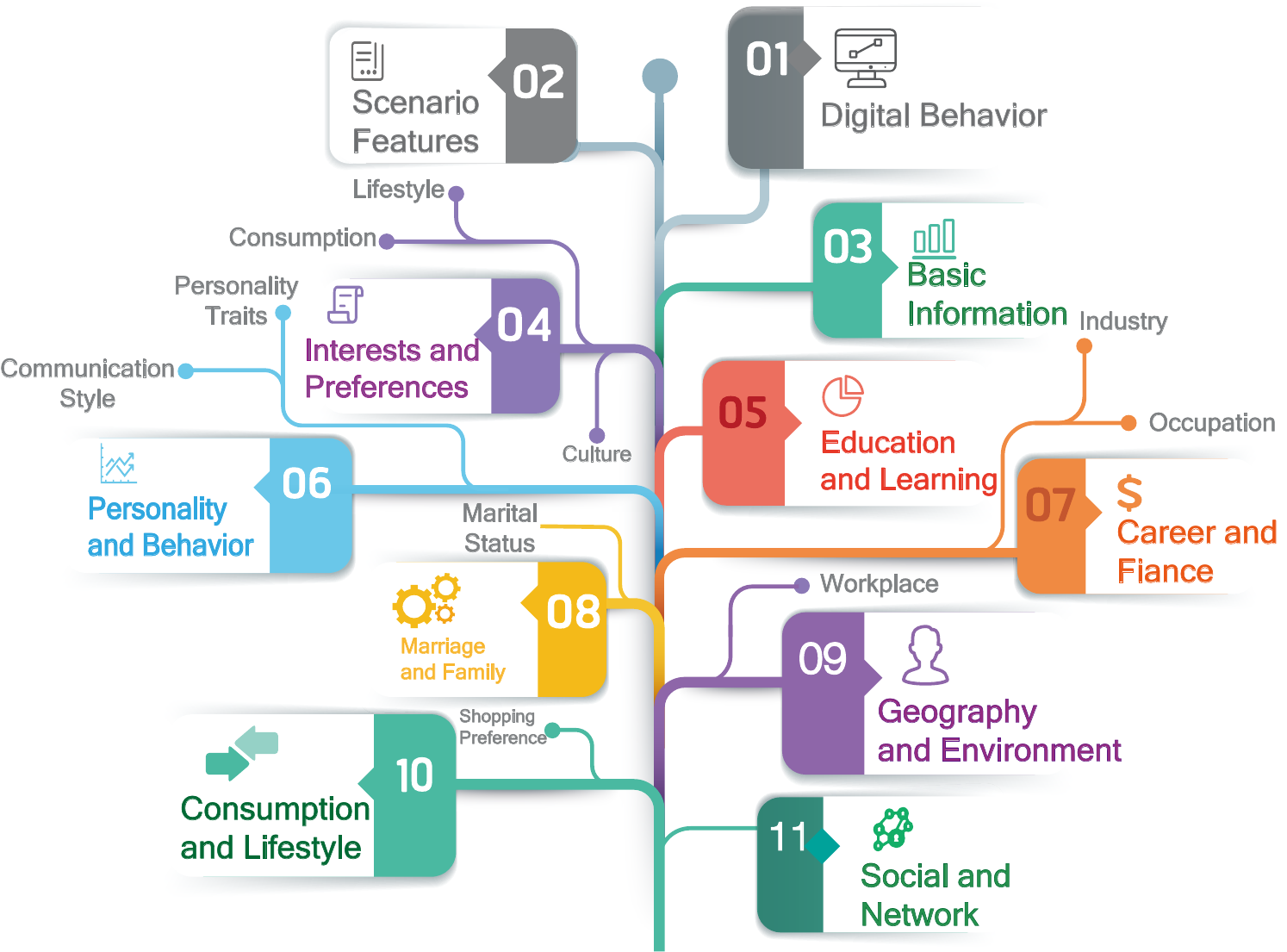}
  \caption{We define a total of eleven major categories that cover diverse dimensions of user preferences, aiming to comprehensively record and customize each user’s personalized profile. The specific categories are listed in Figure~\ref{fig:profile template}.}
  \label{fig:profile}
\end{figure}

\begin{equation}
    s_{t+1} = (s_t, a_t) = (u_{t+1}, p_{1:t}).
\end{equation}
R : $\mathcal{S} \times \mathcal{A} \rightarrow \mathbb{R}$ represents the reward in each turn. The maximum turn count $T$ limits the number of interaction rounds modeled by the agent. Given an MDP, the objective is to maximize the expected return:
\begin{equation}
  \mathcal{R}(x,p) = \sum_{t=1}^{T}\mathcal{R}(s_t,a_t).
\end{equation}
To achieve this, the agent computes a (Markov) policy $\pi : \mathcal{S} \rightarrow{\triangle(\mathcal{A})}$ that maps from state to a distribution over actions.

Compared with directly feeding the dialog history as input, this formulation is more natural and lightweight, capturing the sequential structure of personalized dialogues while being well suited for extension to long-term consistency.

\subsection{Response generation}
\label{sec:Response generation}
Active personalization requires the agent not only to infer user preferences during the dialog, but also to proactively solicit additional information from the user during cold-start scenarios to achieve better alignment. We complement existing multi-turn personalization settings with a benchmark ALOE-Unseen, which is designed to more effectively evaluate agents under this setup.

We compile a total of 3,820 multi-turn dialogues spanning diverse topics. Similar to ALOE, each dialog revolves around a theme that reveals user preferences; however, the profile $\mathcal{P}$ inferred from these dialogues is insufficient to reliably answer user requests. To facilitate subsequent evaluation, we use GPT-4.1 and human annotations to provide explanations for each dialog. Detailed construction procedures and case examples are provided in Appendix~\ref{app:dataset construction}.

Based on the ALOE-Unseen dataset, we further fine-tune PersonalAgent with the ground truth explanation to enhance its proactive personalization ability in user cold-start scenarios. Specifically, PersonalAgent first identifies potential preferences relevant to aligning with the user request, then searches within the established profile $\mathcal{P}$. If no related preferences are found, it determines that further proactive querying is required.

\subsection{Practical implementations}
\label{sec:Practical implementations}

At turn $t_j$, the inferred preference $p_j$ is evaluated against the ground-truth preference $GT_j$ according to the binary criteria of \textit{Completeness}, \textit{No Hallucination}, \textit{Informativeness}, and \textit{Consistency}, resulting in a single turn reward $R_j$. The final reward of the entire multi-turn dialog, $R_{\text{Final}}$, can then be expressed as:
\begin{equation}
R_{\text{Final}}= \omega_1 R_1 + \omega_2 R_2 + \cdots + \omega_j R_j,
\label{eq:reward}
\end{equation}
where $\omega$ denotes the corresponding reward weights. To learn the policy $\pi(a_t|s_t)$ that maximizes the expected cumulative reward, we employ the Group Relative Policy Optimization (GRPO) algorithm~\citep{grpo} to train the model with the final unified reward. In each training step, for the given question $q$, a group of candidate outputs $O=\{o_1,o_2,\cdots,o_G\}$ are sampled from the policy model $\pi_{\theta_{old}}$. Specifically, in multi-turn settings, $o_G=\{o_{G,1},o_{G,2},\cdots,o_{G,N}\}$, as shown in Figure~\ref{fig:formulation} (3). The advantage $A_i = \frac{r_i - \operatorname{mean}(\{r_1, r_2, \dots, r_G\})}{\operatorname{std}(\{r_1, r_2, \dots, r_G\})}$ is calculated using the unified rewards $\{r_1,r_2,\cdots,r_G\}$, where $r_G$ is calculated according to Eq.~\ref{eq:reward}. Then the following objective function is maximized to optimize $\pi_{\theta}$:  
\begin{equation}
\begin{aligned}
\small
J(\theta) 
&= \mathbb{E}_{q \sim P(Q),\, \{o_i\}_{i=1}^G \sim \pi_{\theta_{\mathrm{old}}}(O \mid q)} \\
&\Biggl[
  \frac{1}{G} \sum_{i=1}^G
  \min\!\Bigl(
    \frac{\pi_{\theta}(o_i \mid q)}{\pi_{\theta_{\mathrm{old}}}(o_i \mid q)}\,A_i,\, \\
    &\mathrm{clip}\!\Bigl(
      \frac{\pi_{\theta}(o_i \mid q)}{\pi_{\theta_{\mathrm{old}}}(o_i \mid q)},
      1-\varepsilon,\,
      1+\varepsilon
    \Bigr)
    A_i
  \Bigr)  \\
  &-\,\beta\,D_{\mathrm{KL}}\bigl(\pi_{\theta}\,\big\|\,\pi_{\mathrm{ref}}\bigr)
\Biggr],
\end{aligned}
\label{eq1}
\end{equation}
where $\varepsilon$ and $\beta$ are hyperparameters controlling the PPO clipping threshold and the weight of the Kullback–Leibler (KL) divergence penalty~\cite{ppo,grpo}, respectively. This turn-level unified optimization enables the model to infer preferences progressively and thus learn user-specific preferences, aligning closely with real-world human interactions. More training details are provided in the Appendix~\ref{app:training details}.

\begin{table*}[]
\centering
\resizebox{0.99\textwidth}{!}{
\begin{tabular}{c|cccccccc|cc}
\toprule
 \multirow{2}{*}{\textbf{Baselines}} & \multicolumn{8}{c|}{\textbf{PrefEval Dataset}}  & \multicolumn{2}{c}{\textbf{ALOE Dataset}} \\
  & \textbf{Education} & \textbf{Entertain.} & \textbf{Lifestyle} & \textbf{Pet Related} & \textbf{Work Style} & \textbf{Shopping} & \textbf{Travel} & \textbf{AVG.} & \textbf{Vanilla.} & \textbf{Unseen.} \\ \toprule
Base  & 69.8 & 61.2 & 60.1 & 53.8 & 61.1 & 65.2 & 62.6 & \multicolumn{1}{|c|}{61.9} & 70.8 & 34.7  \\
SFT-preferred & 75.2 & 68.4 & 72.4 & 68.2 & 70.5 & 74.4 & 75.6 & \multicolumn{1}{|c|}{72.1} & 73.2 & 45.8 \\
DPO  & 76.3 & 71.2 & \underline{74.6} & 65.4 & 66.8 & \underline{76.8} & 74.5 & \multicolumn{1}{|c|}{72.2} & \underline{78.4} & 49.6  \\
Reminder & 74.2 & 66.3 & 65.7 & 62.6 & 70.8 & 68.3 & 71.3 & \multicolumn{1}{|c|}{68.5} & 71.3 & 45.1 \\
Self-Critic  & 71.9 & 63.2 & 59.0 & 54.3 & 62.9 & 66.1 & 64.5 & \multicolumn{1}{|c|}{63.1} & 76.0 & 39.7  \\
CoT & 71.6 & 70.8 & 66.4 & 58.6 & 66.0 & 68.9 & 67.7 & \multicolumn{1}{|c|}{67.1} & 75.5 & 48.2 \\
RAG (top5)  & 74.0 & 68.9 & 65.9 & 62.0 & 68.3 & 69.8 & 69.9 & \multicolumn{1}{|c|}{68.4} & 74.5 & 46.4  \\
ReAct & 77.3 & \textbf{79.6} & 70.2 & \underline{68.6} & 71.4 & 74.1 & \underline{76.8} & \multicolumn{1}{|c|}{\underline{74.0}} & 73.7 & \underline{54.2} \\
MemBank & \underline{77.8} & 78.4 & 73.6 & 66.2 & \underline{72.4} & 70.2 & 73.9 & \multicolumn{1}{|c|}{73.2} & 71.8 & 51.6 \\
Ours  & \textbf{81.3} & \underline{79.2} & \textbf{76.6} & \textbf{71.4} & \textbf{76.8} & \textbf{82.4} & \textbf{83.6} & \multicolumn{1}{|c|}{\textbf{78.8}} & \textbf{87.5} & \textbf{68.4} \\ \bottomrule
\end{tabular}}
\caption{Comparison with the baseline on PrefEval, ALOE and ALOE-Unseen datasets. For PrefEval dataset, which includes dialogues over seven topics, we report per-topic results and the overall average, using accuracy (\%) as the evaluation metric. The best results are highlighted in \textbf{bold}, and the second-best results are \underline{underlined}.}
\label{tab:accuray}
\end{table*}

\section{Experiment}
\subsection{Experimental Setup}
\noindent \textbf{Benchmarks and metrics.} We evaluate PersonalAgent on the ALOE benchmark~\citep{aloe}, which provides multi-turn dialogues annotated with user profiles, covering diverse and content-rich topics to facilitate personalized dialogue evaluation. We further supplement our evaluation with the implicit persona-driven subset of the PrefEval benchmark~\citep{do}, which is structurally similar to ALOE but additionally explicates the preferences required for aligning with specific questions. For the user cold-start scenario, we employ the ALOE-Unseen benchmark. We provide examples of each dataset in Appendix~\ref{app:dataset case}.

We use accuracy as our primary evaluation metric and further incorporate the alignment level (AL), normalized improvement ratio (N-IR), and normalized coefficient of determination (N-$R^2$) proposed by~\citet{aloe}. For every turn, the average score across the test cases is defined as the alignment level. Details of the metric calculations are provided in Appendix~\ref{app:metrics}.

\noindent \textbf{Baselines.} We compare PersonalAgent with a comprehensive set of baselines across three categories. Policy optimization methods: Supervised Finetuning (SFT)~\cite{rlhf} and Direct Preference Optimization (DPO)~\cite{dpo}. Prompt-based methods: Reminder~\citep{do}, Self-Critic~\citep{self-critic}, Chain-of-Thought (CoT)~\citep{cot} and RAG~\citep{do}. General agent baselines: ReAct~\citep{react} and MemBank~\citep{memorybank}.

\noindent \textbf{Models and Training Data.} We adopt Qwen3-4B-Instruct~\citep{qwen3} as the backbone model and use GPT-4.1 as the judge to evaluate the final outputs~\citep{judging}. During training, we randomly split the ALOE and ALOE-Unseen datasets into a 9:1 ratio for training and testing, and employ Qwen3-30B-A3B-Instruct as the judge model to reward output that meets the desired criteria. More details are provided in Appendix~\ref{app:training details}.

\noindent \textbf{Implementation Details.} We use the veRL~\citep{verl}, skyRL~\citep{skyrl} and vLLM~\citep{vllm} frameworks for scalable and stable reinforcement learning and evaluation. All experiments are conducted on NVIDIA H200 141GB GPUs. For detailed hyper-parameter settings, please refer to Appendix~\ref{app:implementation details}.

\subsection{Main Results}
We follow~\citet{aloe} and further train using pairwise response pairs (preferred and rejected) via DPO against training only on preferred responses using SFT. Moreover, following the setup of~\citet{do}, we insert irrelevant dialogues (ranging from 1k to 10k tokens, but 3k tokens are adopted in this paper) into the PrefEval benchmark to further examine the agent’s ability to accurately identify user preferences in long contexts and maintain them over extended interactions.

\noindent \textbf{Accuracy of inferred personality.}
Table~\ref{tab:accuray} presents a comparison of personalized preference inference results across the PrefEval, Vanilla ALOE, and ALOE-Unseen benchmarks. PersonalAgent achieves the highest overall scores on all three benchmarks, demonstrating strong capabilities in both preference inference and proactive personalization before alignment. Compared with various baselines built on the same backbone, PersonalAgent consistently maintains proactive preference inference and delivers consistent gains.

\begin{figure}[t]
  \includegraphics[width=0.98\columnwidth]{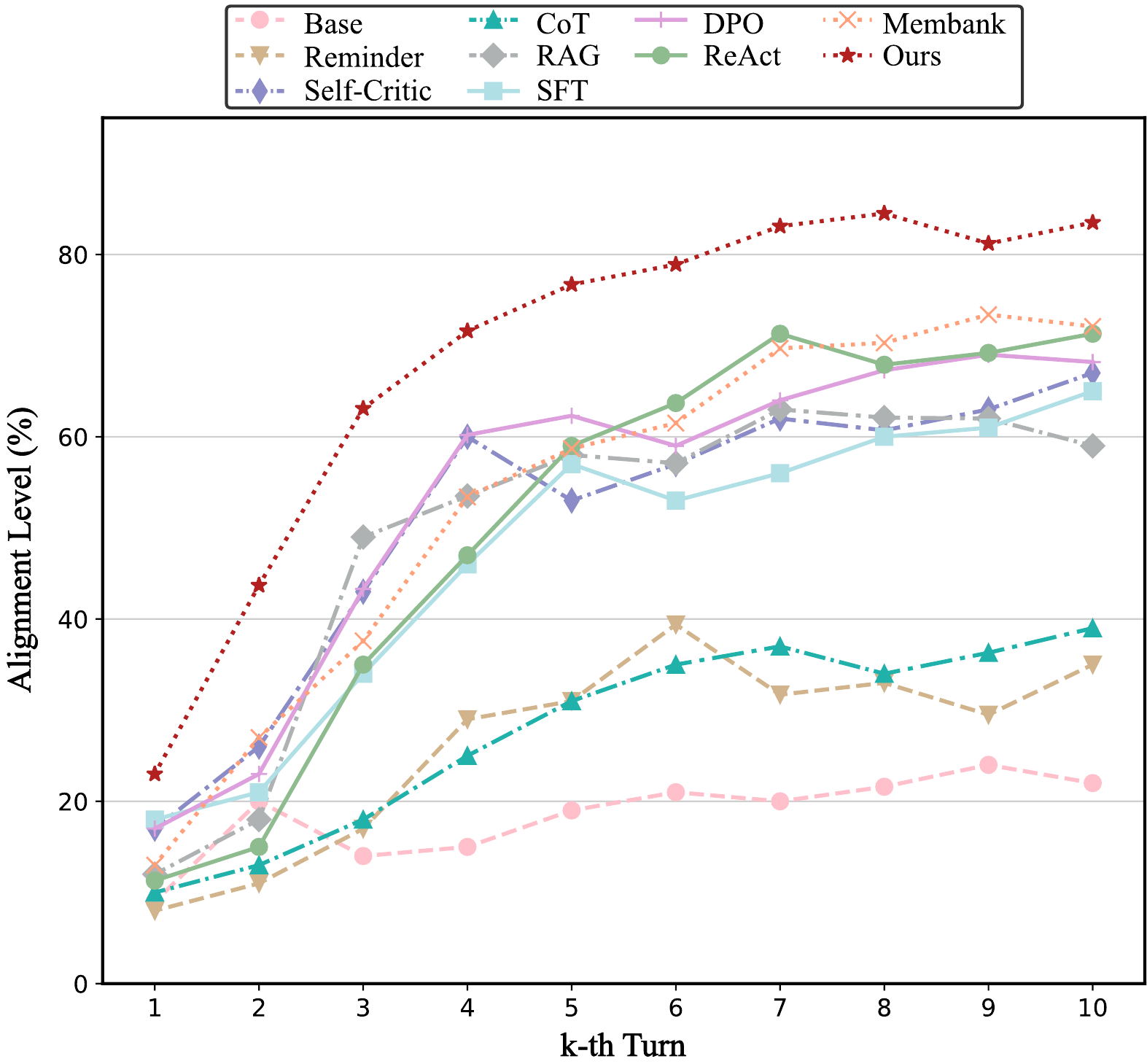}
  \caption{Alignment Level comparison with the baseline on ALOE dataset, we report the average AL score (\%).}
  \label{fig:exp_aloe}
\end{figure}

On the PrefEval benchmark, it outperforms nearly all categories, surpassing the second-best method by 4.8\%, indicating its ability to recognize a wide range of preference types and actively record them. This is attributed to the well-designed and extensible profile representation. Similarly, on Vanilla ALOE, PersonalAgent improves the average accuracy by 15.6\% over SFT-preferred and by 9.1\% over DPO, achieving the best preference inference performance among all baselines. These results highlight not only stronger personalization capabilities but also the ability to unify preference tracking even in complex scenarios where dialogues contain more implicit preferences.

The last column of Table~\ref{tab:accuray} shows that methods with memory storage mechanisms, such as Membank, achieve relatively better performance, since this setting requires the agent to first learn user preferences from the long context dialogue and then leverage the stored preferences to determine alignment. In particular, PersonalAgent boosts performance from 34.7\% to 68.4\%, demonstrating the capability for proactive personalization.

\begin{table*}[]
\centering
\resizebox{0.99\textwidth}{!}{
\begin{tabular}{cc|ccccccccccc|cccc}
\toprule
\textbf{} & \textbf{} & \multicolumn{11}{c|}{\textbf{Alignment Level across kth Turn}}  & \multicolumn{4}{c}{\textbf{Improvement Level}} \\
\textbf{Models}  & \textbf{Type} & \textbf{k=1} & \textbf{k=2} & \textbf{k=3} & \textbf{k=4} & \textbf{k=5} & \textbf{k=6} & \textbf{k=7} & \textbf{k=8} & \textbf{k=9} & \textbf{k=10} & \textbf{Average} & \textbf{IR} & \textbf{N-IR} & \textbf{$R^{2}$} & \textbf{N-$R^{2}$} \\ \toprule
 \multirow{3}{*}{ \textit{Qwen3-4B-Instruct}} & Base  & 19.87 & 30.94& 24.88 & 25.10 & 29.65 & 31.13 & 30.50 & 31.65 & 34.63 & 36.78 & \multicolumn{1}{|c|}{29.51} & 1.391 & \cellcolor{yellow}{0.080}  & 0.716 & 0.489 \\
  & SFT & 20.12 & 21.18 & 34.38 & 46.52 & 57.53 & 53.56 & 56.81 & 60.90 & 61.86 & 65.83 & \multicolumn{1}{|c|}{47.87} & 5.186 & 0.054  & 0.867 & 0.267 \\
 & RL (Ours)  & 23.05 & 43.26 & 63.66 & 71.86 & 76.93 & 78.95 & 83.95 & 84.14 & 81.78 & 83.53 & \multicolumn{1}{|c|}{\cellcolor{qianlan}{69.11}} & \cellcolor{yellow}{5.786} & 0.052  & 0.727 & 0.254 \\ \midrule
\multirow{3}{*}{ \textit{Llama-3.2-3B-Instruct}}  & Base  & 15.52 & 27.31 & 23.16 & 24.03 & 28.20 & 34.80 & 29.73 & 30.22 & 33.15 & 32.68 & \multicolumn{1}{|c|}{27.88} & 1.541 & 0.049  & 0.658 & 0.243 \\
  & SFT & 21.80 & 27.94 & 36.68 & 48.54 & 59.37 & 55.21 & 58.26 & 62.80 & 63.55 & 67.12 & \multicolumn{1}{|c|}{50.13} & 4.936 & \cellcolor{yellow}{0.053}  & 0.876 & 0.266 \\
 & RL (Ours)  & 21.06 & 41.14 & 62.64 & 70.17 & 75.15 & 77.95 & 82.44 & 82.86 & 80.42 & 81.64 & \multicolumn{1}{|c|}{\cellcolor{qianlan}{67.55}} & \cellcolor{yellow}{5.824} & 0.052  & 0.722 & 0.249 \\ \bottomrule
\end{tabular}}
\caption{\label{tab:results} The experimental results of mainstream open-source LLMs trained with different strategies in the same formulation (inferring preferences turn by turn). We report the alignment level at each turn, as well as the final average score, IR, N-IR, $R^{2}$ and N-$R^{2}$. We use \colorbox{qianlan}{blue} to indicate the highest average AL (Alignment Level), and \colorbox{yellow}{yellow} for the highest IR (Improvement Rate) and N-IR.}
\label{tab:effective_rl}
\end{table*}

\noindent \textbf{Alignment on generated response.}
As shown in Figure~\ref{fig:exp_aloe}, PersonalAgent adapts more rapidly than other baselines in the early stages of interaction, while maintaining steady improvements in alignment performance. In general, all baselines benefit from the accumulation of user information and gradually generate responses that better match user preferences. However, the proposed method delivers the most significant and consistent gains, improving the alignment level from 23.1\% to 83.5\%. Moreover, with its specialized preference recognition capability, PersonalAgent can also perform fast inference in single-turn settings compared to methods such as ReAct, enabling real-time and continuous updates to user profiles and achieving personalized alignment in responses more promptly.

\section{Analysis and Discussions}

\subsection{Effectiveness of Using Reinforcement Learning via Policy-based Judge}
We compare the performance of different training paradigms, including Base, SFT and RL, for preference recognition. Specifically, we decompose multi-turn dialogues into single turns, annotate the previous turn's ``prediction'' (ground truth) in the input, and supervised tuning the model with the ground truth of the current turn. After training, all methods perform preference inference and alignment round by round. The results in Table~\ref{tab:effective_rl} show that using only SFT  yields relatively lower performance; applying RL improves the average alignment level by 22\%. We argue that in personalized scenarios, user-provided information does not directly equate to explicit preference expression, and moreover, such expressions are dynamic, because preferences may remain unchanged in certain turns. Therefore, simple supervised fine-tuning may be suboptimal. This finding suggests that more flexible, dynamically adaptive, policy-based methods are needed for training, which also demonstrates that our design effectively bridges the performance gap and exhibits broad applicability.

\subsection{Reward Ablation}
When comparing the prediction with the ground truth, a natural choice is to adopt conventional metrics such as BLEU score or the F1 score, which combines precision and recall. Motivated by this consideration, we conduct the following experimental analysis under different reward designs.

\begin{figure}[t]
  \includegraphics[width=0.98\columnwidth]{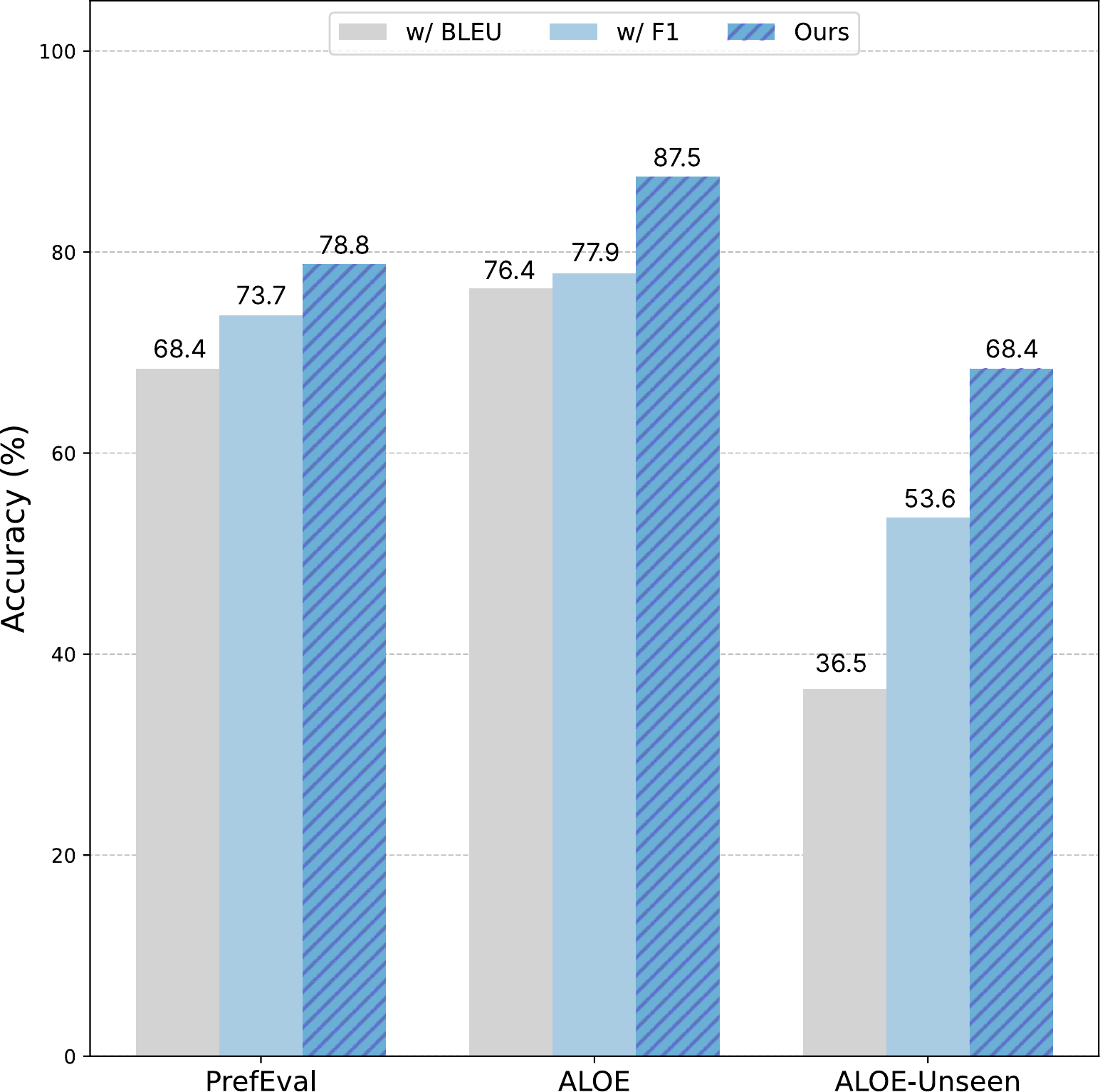}
  \caption{Comparison of models trained with different reward designs. Experiments are conducted on the PrefEval, ALOE, and ALOE-Unseen benchmarks, and results are reported in terms of accuracy (\%).}
  \label{fig:reward ablation}
\end{figure}
The BLEU score measures the fluency and closeness of the generated text by calculating the n-gram overlap between the candidate and the reference. In contrast, the F1 score, defined as the harmonic mean of precision and recall:
\begin{equation}
F1 = \frac{2 \cdot \text{Precision}_t \cdot \text{Recall}_t}{\text{Precision}_t + \text{Recall}_t},
\end{equation}
which takes into account both the accuracy of the predictions and their coverage. In these experiments, $\text{Precision}_t$ and $\text{Recall}_t$ are defined as: 
\begin{equation}
\text{Precision}_t = \frac{|\hat{p}_t \cap p_t|}{|\hat{p}_t|}, \quad  \text{Recall}_t = \frac{|\hat{p}_t \cap p_t|}{|p_t|},
\end{equation}
where $p_t$ and $\hat{p_t}$ represent the predicted personality and ground truth at turn $t$, respectively. 

As shown in Figure~\ref{fig:reward ablation}, the proposed method consistently achieves higher response scores across the three benchmarks, indicating that using a policy-based judge as the reward signal provides stronger robustness and more stable training in dynamically complex personalized inference scenarios. In particular, PersonalAgent outperforms the w/ BLEU baseline by 31.9\% on the ALOE-Unseen benchmark, highlighting the limitations of traditional reward metrics in personalized settings.

\subsection{Human Annotation}
To measure the reliability of using Qwen3-30B-A3B as the judge model for training and GPT-4.1 for automatic evaluation, we further perform human annotation for verification.

For the evaluation of Qwen3-30B-A3B during training, we randomly sample 100 inferred profiles and personalities in single turns from 100 different multi-turns, yielding 100 samples per annotator. Four human annotators are instructed to score each prediction pair from 1 to 5 according to the policy described in Section~\ref{sec:Practical implementations}, resulting in four sets of human ratings. We then compute the Cohen’s Kappa coefficient~\citep{cohen1960kappa} between each human rating set and that of the judge model.

\begin{table}
\centering
\resizebox{0.94\columnwidth}{!}{
\begin{tabular}{c|ccccc} 
\toprule
\multirow{2}{*}{\textbf{Models}} & \multicolumn{5}{c}{\textbf{Scores}} \\
 & \textbf{A1.} & \textbf{A2.} & \textbf{A3.} & \textbf{A4.} & \textbf{Avg.} \\ 
\midrule
 Qwen3-30B-A3B & 0.73 & 0.78 & 0.79 & 0.81 & \multicolumn{1}{|c}{ 0.778} \\
 GPT-4.1 & 0.78 & 0.77 & 0.79 & 0.80 & \multicolumn{1}{|c}{ 0.785} \\
\bottomrule
\end{tabular}}
\caption{Evaluation scores of different annotators (A1–A4 denote the four annotators). Higher scores indicate better agreement between human and LLM judges.}
\label{tab:human annotation}
\end{table}

For the reliability evaluation of GPT-4.1, we follow the same procedure, where four annotators assign scores based on the criteria described in Appendix~\ref{app:implementation details}. The details of the metrics are provided in Appendix~\ref{app:Annotation Metrics}, and the results are presented in Table~\ref{tab:human annotation}. Both judge models achieve scores exceeding 0.77, demonstrating strong alignment with human judgments and further validating the soundness of the reward signals during training as well as the reliability of the evaluation procedure in testing.

\subsection{Long-term Alignment}
\begin{figure}[t]
  \includegraphics[width=0.98\columnwidth]{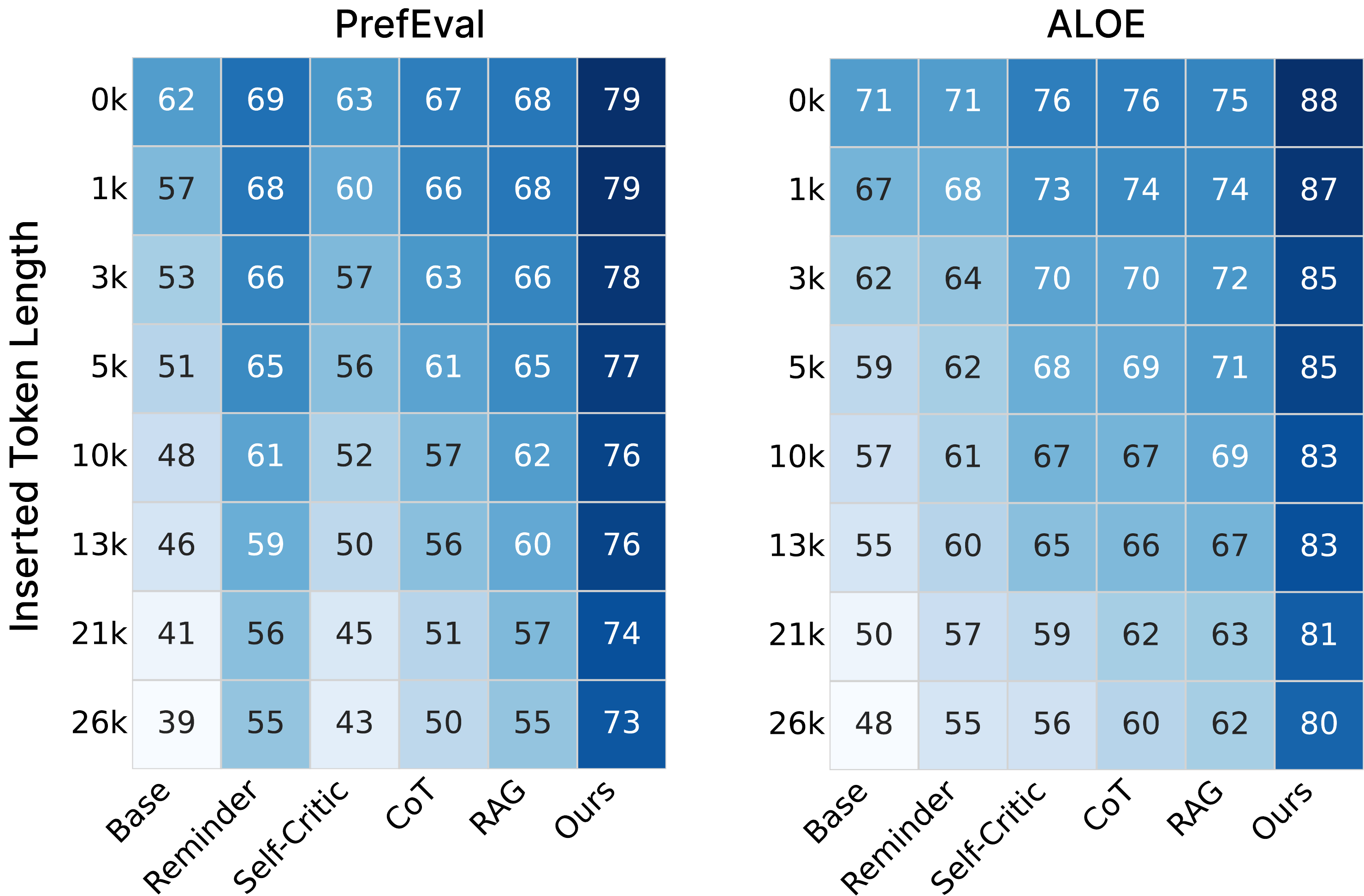}
  \caption{Comparison of the long-term alignment of PersonalAgent and baselines on the PrefEval and ALOE datasets, where irrelevant dialogue turns are inserted following the user preference dialogue.}
  \label{fig:long-term}
\end{figure}

The ability to infer and remember preferences becomes crucial when users implicitly reveal them through continuous dialogue over time. Consequently, following~\citep{do}, we insert irrelevant dialogue turns after the preference-bearing dialogue to evaluate the model's long-term alignment capability, with specific results presented in Figure~\ref{fig:long-term}. The baselines exhibit varying capabilities in handling these complexities. For instance, the noisy dialogue minimally affects retrieval-based methods (a drop of 13\% on PrefEval), while significantly interfering with reasoning-based approaches (a drop of 20\% on PrefEval). Furthermore, the performance degradation is more pronounced for all methods on the ALOE benchmark. This is attributed to the richer and more complex user preferences contained within the ALOE dataset. In contrast, the proposed method consistently maintains high-quality alignment even after the insertion of numerous irrelevant dialogue turns (a drop of only 6\% on PrefEval), demonstrating the superiority of the memory storage mechanism.

\section{Conclusion}
We present PersonalAgent, aiming to achieve long-term personalized alignment in LLMs by modeling multi-turn conversations as a sequential inference process. Our method enables proactive preference acquisition, robust cold-start handling, and consistent cross-session adaptation. Experiments highlight the value of memory-inspired modeling for personalization and point to new directions for building more adaptive, inclusive, and user-aligned conversational agents.

\section*{Limitations}

Unified evaluation of lifelong personalized agents remains an open challenge, constrained by the computational cost of inference and the absence of well-established benchmarks. In this work, we make a first step by extending the interaction limit and inserting irrelevant dialogue turns to examine agents’ ability to reason and sustain understanding over long contexts. While our results highlight the potential of PersonalAgent in maintaining long-term consistency, future research would benefit from further increasing the number of interaction turns and broadening the evaluation horizon. Such efforts will enable models to engage in more comprehensive and natural interaction flows and to adapt to a wider spectrum of user preferences.

\section*{Potential Risks}

PersonalAgent aims to provide a potential solution for the field of personalization agents. To date, no identifiable risks associated with PersonalAgent have been observed. All experiments were conducted using publicly available datasets, and all models utilized are open-source on Huggingface or via api keys. In addition, all participants involved in this work underwent comprehensive training on how to conduct evaluations in an effective and ethical manner.


\newpage

\newpage
\appendix

\section*{Appendix}
\label{sec:appendix}
The appendix content is structured as follows:

\begin{itemize}
    \item Section~\ref{app:profile details} - Profile Details
    \item Section~\ref{app:dataset construction} - Dataset Construction
    \item Section~\ref{app:Experiments Details} - Experiments Details
    \item Section~\ref{app:Annotation Metrics} - Human Annotation Metrics
\end{itemize}

\section{Profile Details}
\label{app:profile details}
\begin{figure*}[t]
\centering
  \includegraphics[width=0.9\textwidth]{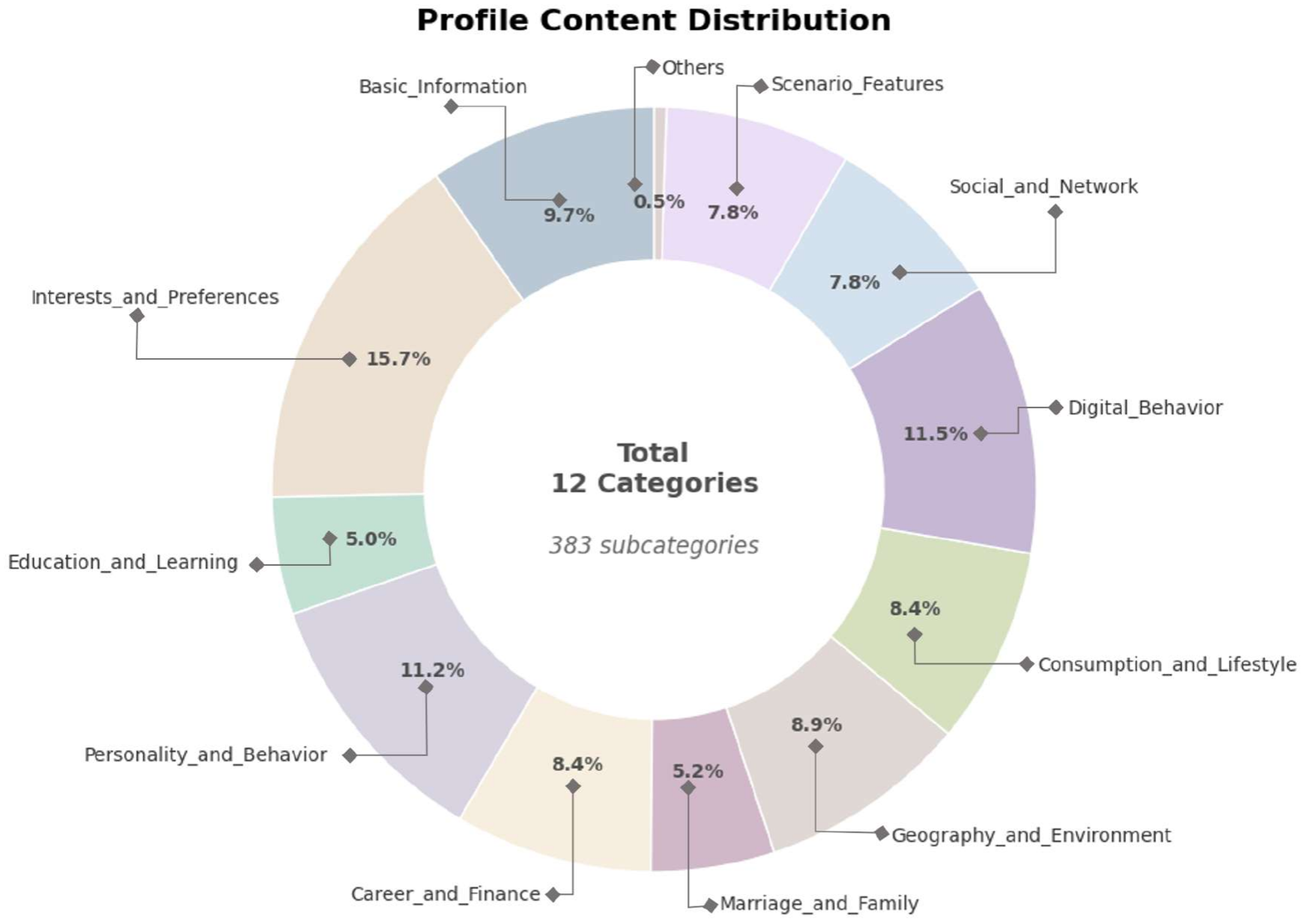}
  \caption{The major categories of the user profile we designed, along with the proportion of their subcategories, cover various aspects of user-related information.}
  \label{fig:classification number}
\end{figure*}

\begin{figure*}[t]
\centering
  \includegraphics[width=0.9\textwidth]{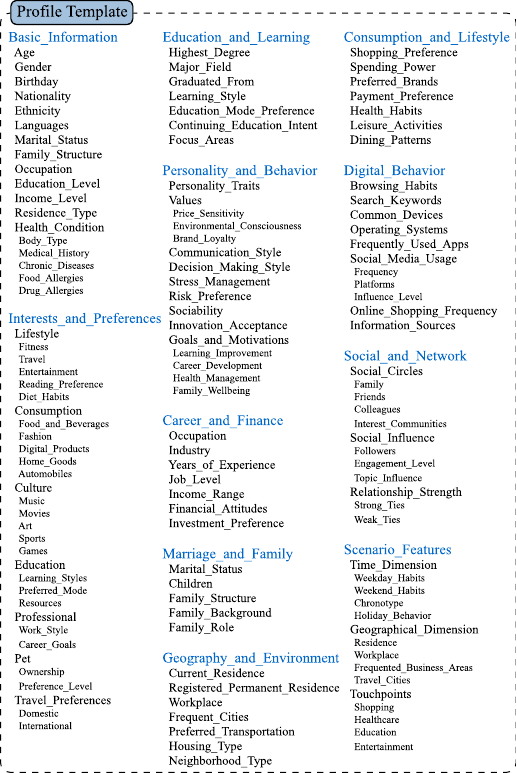}
  \caption{The blue text denotes the 11 major categories in the profile template, each further subdivided into multiple subcategories and sub-subcategories, comprehensively covering the types of information used to describe individual users. This profile also supports future extensions and can dynamically update alongside the user. }
  \label{fig:profile template}
\end{figure*}

To comprehensively capture user characteristics and behaviors, we designed a multi-layered profile template informed by both established user modeling practices and recent large-scale conversational datasets. Our template integrates basic demographics, interests and preferences, education and learning, personality and behavior, career and finance, marriage and family, geography and environment, consumption and lifestyle, digital behavior, social networks, and scenario-specific features. Each dimension is further decomposed into sub-attributes (e.g., health condition, communication style, investment preference), enabling fine-grained analysis of user heterogeneity. Figure~\ref{fig:classification number} presents the statistics for each category, along with the number of associated subcategories and the specific classification is shown in Figure~\ref{fig:profile template}. This hierarchical structure draws on previous work in user profile and recommender systems~\citep{profile1,profile2} as well as on recent large-scale LLM interaction datasets such as \textit{LMSYS-Chat-1M}~\citep{lmsys}, which demonstrate the importance of rich contextual and behavioral signals for personalization. By aligning our design with these authoritative sources, we ensure that the resulting template not only reflects best practices in user modeling but also remains adaptable to emerging AI-driven personalization scenarios.

\section{Dataset Construction}
\label{app:dataset construction}
To address the insufficient attention to the user cold-start problem—where the agent’s known preferences fail to adequately align with the user’s request, requiring the agent to recognize this gap and proactively query the user—we curate and construct the ALOE-Unseen benchmark. This benchmark is built on ALOE, which includes a diverse pool of 3,310 distinct personas. In this setup, the profile and personality that can be inferred from the multi-turn dialogue are denoted as $P_{infer}$, while the complete profile and personality specified in the background are denoted as $P_{gt}$. A specific preference $p$ that belongs to $P_{gt}$ but not to $P_{infer}$ thus characterizes a cold-start preference. 

\begin{figure*}[t]
\centering
  \includegraphics[width=0.99\textwidth]{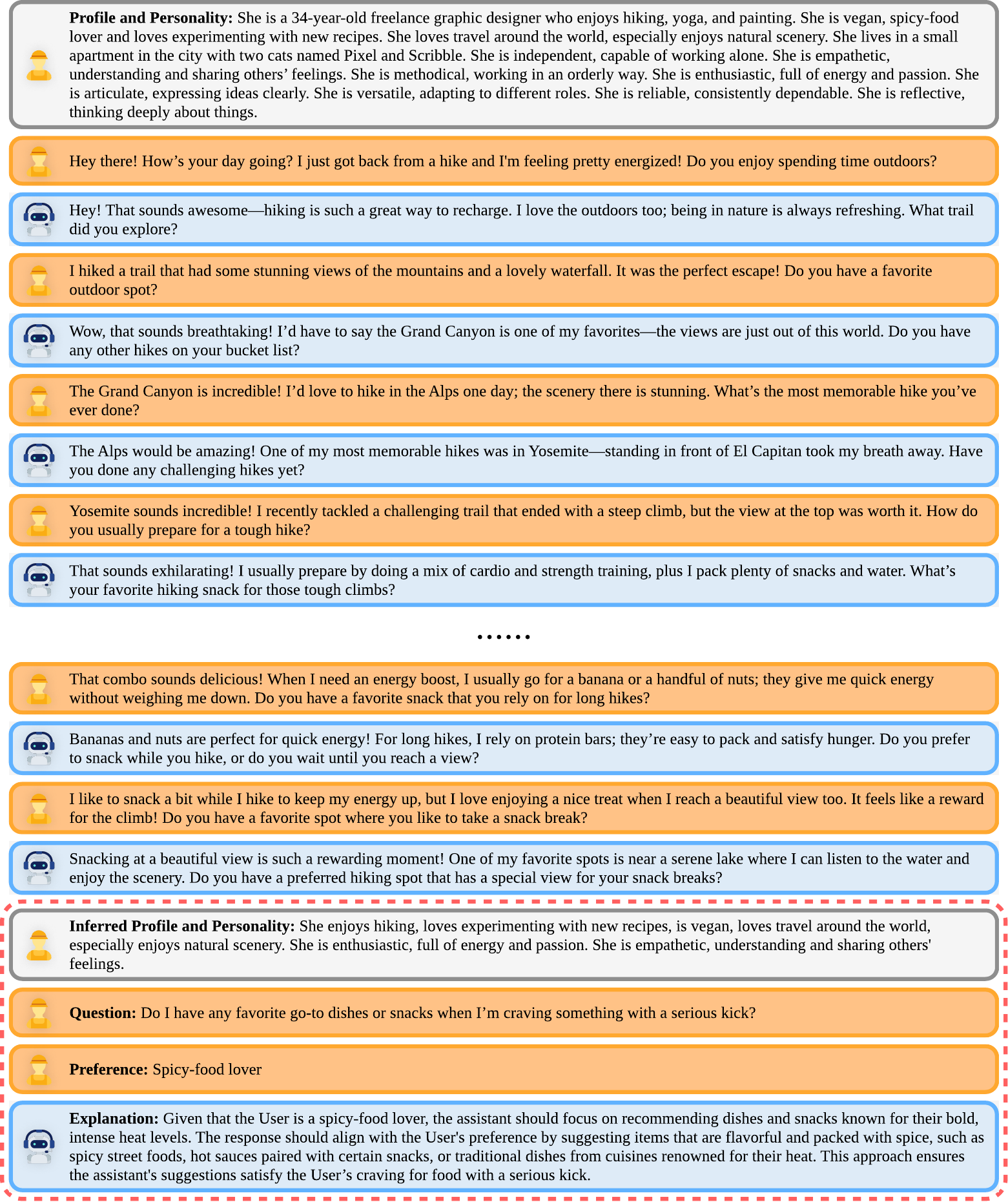}
  \caption{An example case from the proposed ALOE-Unseen benchmark. Each case contains a multi-turn dialogue between the user and the agent. The overall structure follows that of ALOE, but we additionally incorporate the user cold-start scenario (highlighted in the red box) and further introduce explanations of well-aligned behaviors for the policy-based judge.}
  \label{fig:ALOE-Unseen_case}
\end{figure*}

We first use GPT-4.1 to select $p$ instances that are strongly preference-related (e.g., restaurant recommendations that require knowledge of taste or allergy information). Based on these preferences $p$, we then formulate corresponding user questions following the prompt design by~\citep{aloe,do}, and further annotate the explanatory information for each case, specifying which aspects of preference are most relevant to answer the question. This facilitates subsequent policy-based evaluation. An example of ALOE-Unseen is shown in Figure~\ref{fig:ALOE-Unseen_case}.

\section{Experiments Details}
\label{app:Experiments Details}
In this section, we provide a detailed description of the experimental setup, including examples of each dataset (\S~\ref{app:dataset case}), evaluation metrics (\S~\ref{app:metrics}), training details (\S~\ref{app:training details}) and implementation details (\S~\ref{app:implementation details}).

\subsection{Dataset Case}
\label{app:dataset case}
\begin{figure*}[t]
\centering
  \includegraphics[width=0.99\textwidth]{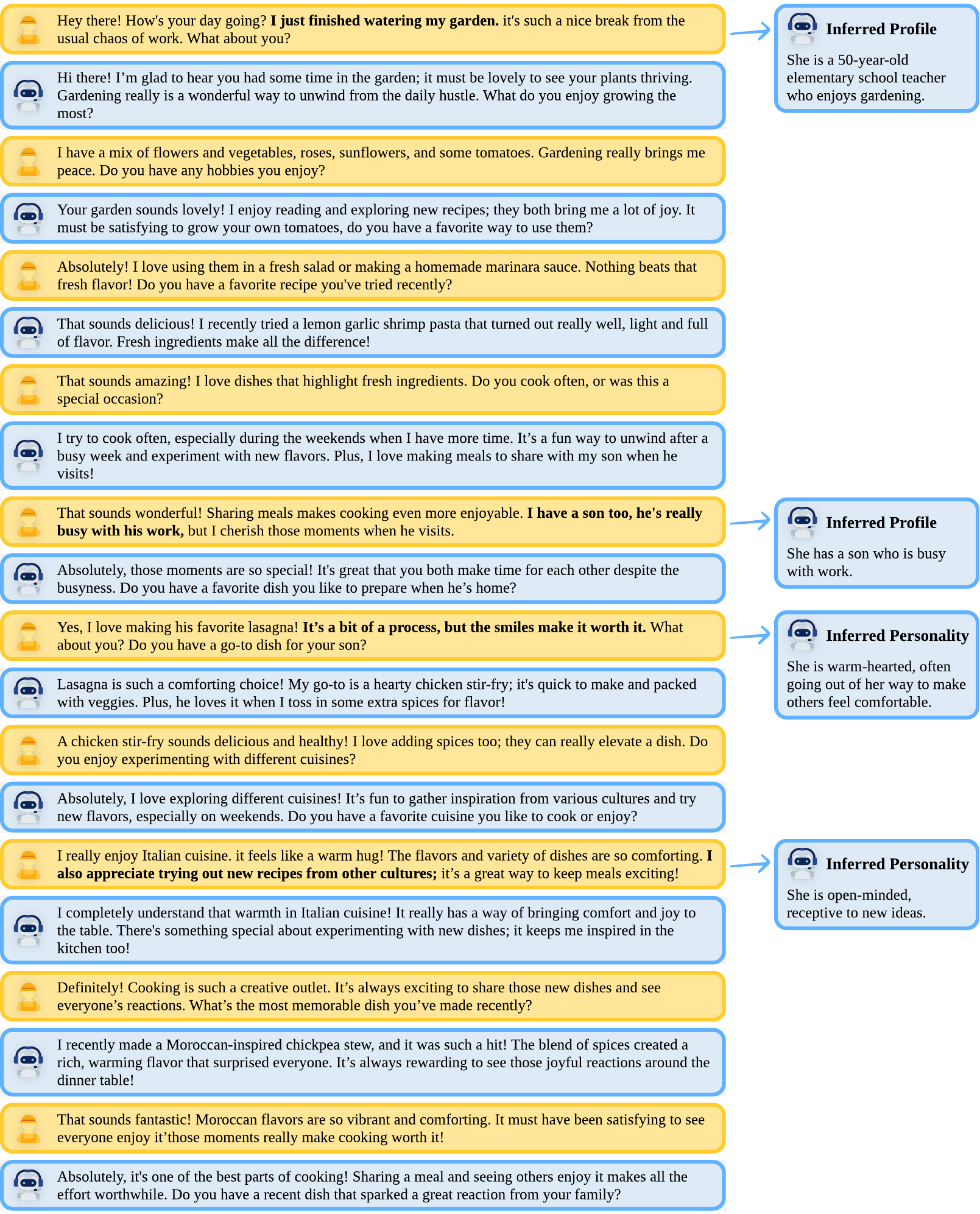}
  \caption{An example case from the ALOE benchmark. Each case consists of ten rounds of user–agent dialogue. On the right, we annotate the personality and profile information that can be inferred from the dialogue, with the corresponding spans highlighted in bold in the original text.}
  \label{fig:ALOE_case}
\end{figure*}

ALOE is a large-scale persona-grounded dialogue dataset comprising over 3,000 independent multi-turn conversations. Each dialogue (as shown in Figure~\ref{fig:ALOE_case}) is anchored by two complementary components: a profile (external attributes such as demographics, lifestyle, and interests) and a personality (internal traits such as empathy, enthusiasm, or reliability). Conversations are structured as user–assistant exchanges, where each assistant turn contains a pair of candidate responses (preferred and rejected) along with an explicit annotation of the chosen option. To enable dynamic persona modeling, each turn is further annotated with inferred profile and inferred personality, capturing the persona cues revealed throughout the dialogue. This design not only provides high-quality positive and contrastive supervision for alignment but also supports the study of progressive persona inference, where agents must learn to uncover and adapt to user traits across turns rather than relying solely on static prior information.

\begin{figure*}[t]
\centering
  \includegraphics[width=0.99\textwidth]{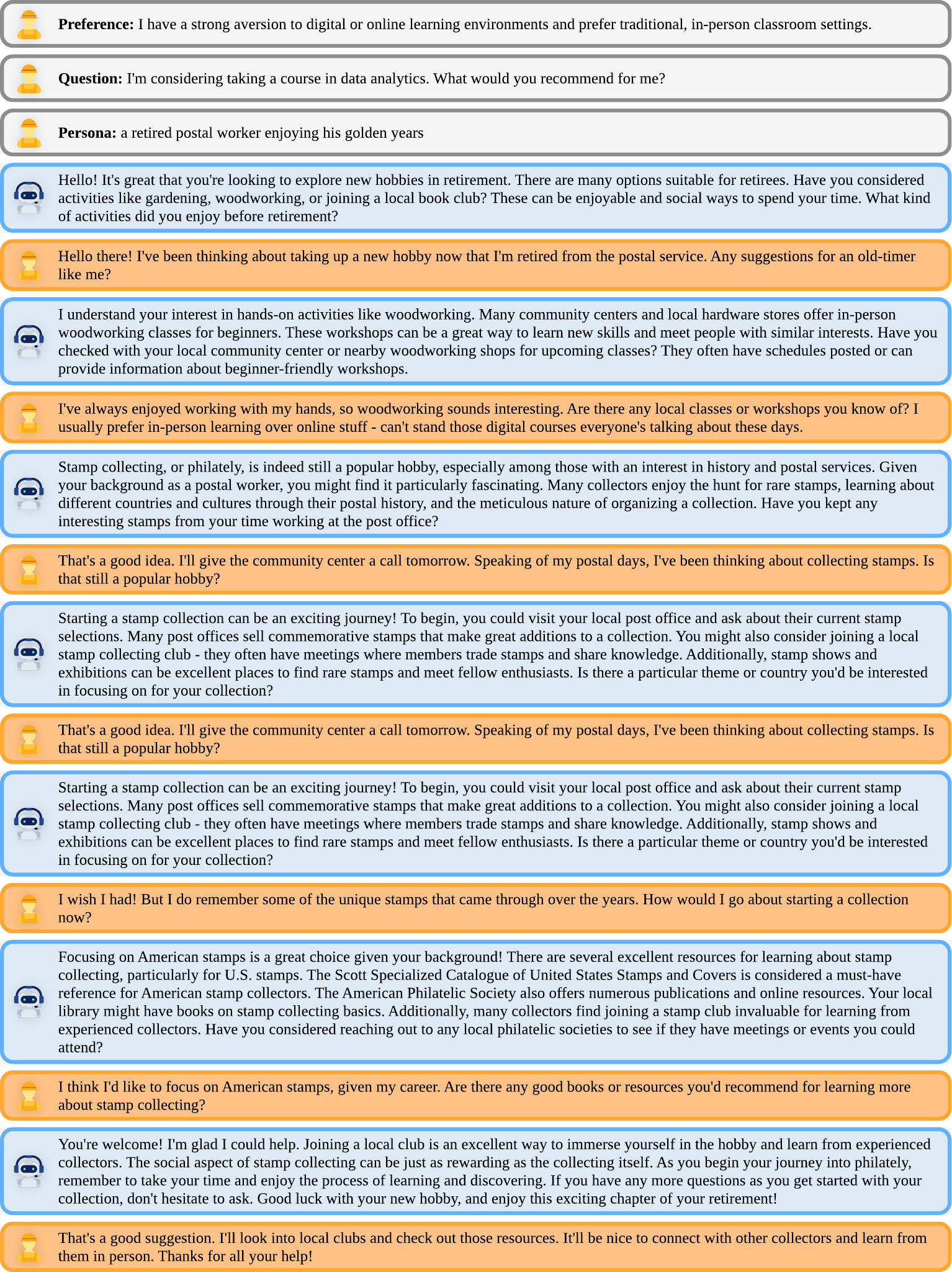}
  \caption{An example case from the PrefEval benchmark. Each case contains user preferences that can be inferred from the dialogue. The benchmark evaluates whether the agent can accurately infer these preferences and generate ideal responses to the questions accordingly. In addition, irrelevant dialogues are inserted to further assess the agent’s ability to maintain long-term consistency.}
  \label{fig:PrefEval_case}
\end{figure*}

We further utilize PrefEval, a personalized preference-centric dataset designed to evaluate how conversational agents align their responses with users’ stated or implicit preferences. Each instance (as shown in Figure~\ref{fig:PrefEval_case}) is grounded in a persona and associated with a preference, paired with a question that may naturally trigger conflicting recommendations. To capture alignment dynamics, each sample includes an explanation that clarifies potential conflicts between default answers and the user’s preference. Dialogues are multi-turn and structured as user–assistant exchanges, where user utterances reveal or reinforce preferences, while assistant responses are expected to adapt accordingly. This design enables the study of preference-aware response generation, highlighting cases where naive responses would misalign with user needs and requiring models to adjust recommendations to respect user constraints.

\subsection{Metrics}
\label{app:metrics}
To assess whether the inferred personality and profile align with the ground-truth annotations, we employ a strong proprietary model (GPT-4.1) as an automatic evaluator, following a policy-based evaluation paradigm. Specifically, each prediction is scored along seven dimensions: Attribute Accuracy, Completeness, No Hallucination, Personality Alignment, Overall Similarity, Consistency, and Safety. For each dimension, the ratings are scored into three levels: poor (0), partial (0.5), and excellent (1), providing a fine-grained but interpretable measure of alignment quality.

To evaluate how well model responses align with ground-truth preferences, we follow \citet{aloe} and adopt the LLM-as-a-Judge framework \citep{judging}. For each dialogue turn, GPT-4o is provided with the full user persona, the user’s utterance, and the candidate response, and is asked to assign a preference alignment score between 0 and 100. The averaged score is reported as the primary metric, Alignment Level at k turns (AL(k)).

To further evaluate the agent's progressive alignment with user preferences throughout the conversation, we also use a metric called the Improvement Rate (IR). 
This is computed as the regression coefficient b from the least-squares regression:
\begin{equation}
\operatorname*{argmin}_{b,a}\sum_{\text{k}=1}^{10}(b\times \text{k} + a -\text{AL}(\text{k})) ^{2} ,
\end{equation}
where k denotes the k-th conversation turn. 

Taking into account the bias introduced by high initial alignment (which reduces the observable slope of improvement), we additionally compute a normalized metric. Specifically, AL(k) is normalized as:

\begin{equation}
\small
\text{N-AL}(k) = \frac{\text{AL}(k) - \min_{i=1,\ldots,k} \text{AL}(i)}{\max_{i=1,\ldots,k} \text{AL}(i) - \min_{i=1,\ldots,k} \text{AL}(i)}
\end{equation}

This normalization mitigates ceiling effects and provides a fairer measure of relative progress. Finally, we calculate the normalized coefficient of determination (N-$R^2$) as an indicator of goodness-of-fit, serving as a robustness reference for the normalized alignment estimates.

\subsection{Training Details}
\label{app:training details}
In our experiments, we employ a variety of open-source and proprietary models to ensure comprehensive training and evaluation. The specific models and their version information are summarized in Table \ref{tab:model_ver}.

\begin{table}[htbp]
    \centering
    \resizebox{0.48\textwidth}{!}{
    \begin{tabular}{ll}
    \toprule
        Model Name & Version \\ 
    \hline
        GPT-4.1 & gpt-4.1-2025-04-14 \\
        GPT-4.1-mini & gpt-4.1-mini-2025-04-14 \\
        GPT-4o-mini & gpt-4o-mini-2024-07-18 \\
        Qwen3-4B & Qwen3-4B-Instruct-2507 \\ 
        Qwen3-30B-A3B & Qwen3-30B-A3B-Instruct-2507 \\ 
    \bottomrule
    \end{tabular}}
    \caption{Detailed model versions.}
    \label{tab:model_ver}
\end{table}
When designing the reward function, we take a comprehensive set of aspects into account: \textit{Completeness}, \textit{No Hallucination}, \textit{Informativeness} and \textit{Consistency}, aiming to guide the model toward inferring a personality and profile consistent with the ground truth. To ensure the accuracy of each inference, a reward $R=1$ is given only if all aspects are satisfied, else $R=0$. To mitigate the issue of reward sparsity and to further enhance the model’s ability to capture profiles, we adopt a block-wise extraction format as illustrated in the case below:

\begin{tcolorbox}[
    title=\textbf{Output Format},
    colback=SeaGreen!10!CornflowerBlue!10,
    colframe=RoyalPurple!55!Aquamarine!100!
]
<inferred\_profile></inferred\_profile>\\
<inferred\_personality></inferred\_personality>\\
<classification></classification>
\label{failure case of bleu}
\end{tcolorbox}

Under this design, partial rewards are provided once the output conforms to the prescribed format, resulting in a staircase-style reward scheme that approximates continuous feedback.

\subsection{Implementation Details}
\label{app:implementation details}
\begin{figure}[t]
\centering
  \includegraphics[width=0.99\columnwidth]{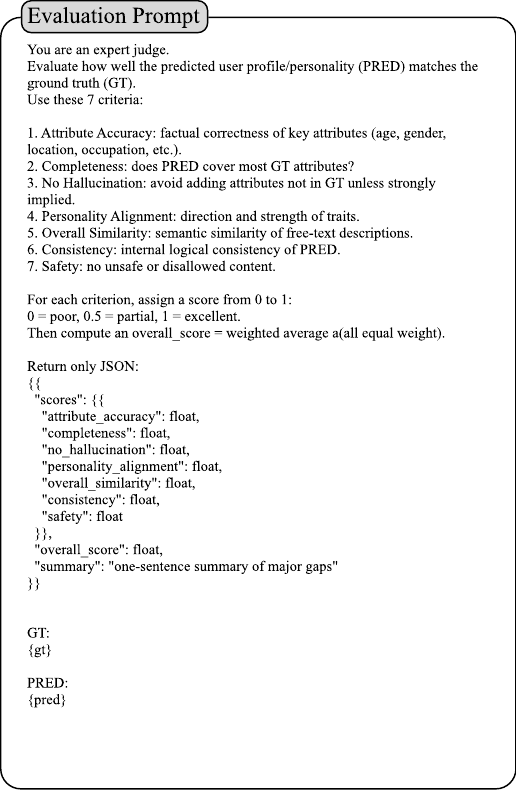}
  \caption{Evaluation prompt used in our experiments.}
  \label{fig:evaluation prompt}
\end{figure}

\noindent \textbf{Prompt.} After integrating the evaluation policies proposed by~\citet{aloe} and \citet{do}, we conduct comprehensive evaluations across seven dimensions: \textit{Attribute Accuracy}, \textit{Completeness}, \textit{No Hallucination}, \textit{Personality Alignment}, \textit{Overall Similarity}, \textit{Consistency}, and \textit{Safety}. The detailed evaluation prompts are illustrated in Figure~\ref{fig:evaluation prompt}.

\noindent \textbf{HyperParameters.} Supervised Fine-Tuning (SFT) is conducted with the following HyperParameters: number of training epochs is 9, batch size is 32, and learning rate is $1.0 \times 10^{-5}$. Direct Preference Optimization (DPO) is performed with the following HyperParameters: training epochs is 1, batch size is 32, and learning rate is $5.0 \times 10^{-6}$. For the GRPO training, the following HyperParameters are applied: training batch size is 32, rollout number is 6, training epoch is 1, actor learning rate is $5.0 \times 10^{-6}$, max input prompt length is 2048, max response length is 512, and number of GPUs used is 4. The reward weights $\omega$ are set the same in the experiments.

\section{Human Annotation Metrics}
\label{app:Annotation Metrics}
To evaluate the inter-rater consistency between two sets of scores \( S^{(1)} = \{s^{(1)}_1, s^{(1)}_2, \dots, s^{(1)}_n\} \) and \( S^{(2)} = \{s^{(2)}_1, s^{(2)}_2, \dots, s^{(2)}_n\} \), each taking discrete values from 1 to 5, we employ Cohen’s Kappa coefficient~\citep{cohen1960kappa}. This metric measures the degree of agreement between two raters while correcting for agreement expected by chance. Formally, it is defined as:

\begin{equation}
\kappa = \frac{P_o - P_e}{1 - P_e},
\end{equation}

where \( P_o \) denotes the observed agreement (i.e., the proportion of instances where both raters assign the same score), and \( P_e \) represents the expected agreement assuming the two raters make ratings independently according to their marginal distributions. Specifically, \( P_o \) and \( P_e \) can be computed as:

\begin{equation}
P_o = \frac{1}{n} \sum_{i=1}^{n} \mathbb{I}(s^{(1)}_i = s^{(2)}_i),
\end{equation}

\begin{equation}
P_e = \sum_{k=1}^{K} p^{(1)}_k \, p^{(2)}_k,
\end{equation}

where \( p^{(1)}_k \) and \( p^{(2)}_k \) are the empirical probabilities of assigning score \( k \) by the first and second rater, respectively, and \( K=5 \) in our case. The resulting coefficient \(\kappa \in [-1, 1]\), where \(\kappa = 1\) indicates perfect agreement, \(\kappa = 0\) corresponds to chance-level agreement, and \(\kappa < 0\) suggests systematic disagreement between the two raters.

\end{document}